\documentclass{article}
\usepackage{spconf,amsmath,graphicx,hyperref}

\usepackage{booktabs}
\usepackage{xcolor}
\usepackage{multirow}

\usepackage{makecell}
\usepackage{pifont}
\newcommand{\cmark}{\ding{51}} 
\usepackage[numbers,sort&compress]{natbib}
\usepackage{csquotes}

\title{Phoneme-Level Visual Speech Recognition via Point-Visual Fusion and Language Model Reconstruction}
\name{Matthew Kit Khinn Teng, Haibo Zhang, Takeshi Saitoh}
\address{Kyushu Institute of Technology, Japan\\
Kyushu Institute of Technology, Fukuoka 820-8502, Japan\\}
\begin{document}
%
\maketitle
\begin{abstract} 
Visual Automatic Speech Recognition (V-ASR) is a challenging task that involves interpreting spoken language solely from visual information, such as lip movements and speaker-dependent variance. A major difficulty arises from viseme ambiguity, where multiple phonemes share similar visual appearances, making word- or character-level prediction error-prone. To address this, our work focuses on phoneme-level modeling, which provides a finer linguistic granularity than visemes while reducing the reliance on large-scale pretraining. We propose a novel phoneme-based two-stage framework that fuses visual and landmark motion features, followed by a Large Language Model (LLM) for sentence reconstruction to address these challenges. Stage 1 consists of Point Visual Automatic Speech Recognition (PV-ASR), which outputs the predicted phonemes addressing speaker-specific facial characteristics. Stage 2 comprises No Language Left Behind (NLLB) LLM, which reconstructs the output phonemes back to words. Our method demonstrates promising performance in terms of Word Error Rates (WER), especially on LRS2 (16.0\% WER) while maintaining competitive results on LRS3 (20.3\% WER).

\end{abstract}

\section{Introduction}
\label{sec:intro}
Lip-reading, often considered a form of V-ASR, refers to the derivation of spoken words from visual cues such as lip movements and other facial articulatory features. The goal of early lip-reading techniques, such as \cite{prajwal_2022subword,ma_2023avsr,liu_2023synthvsr,djilali_2023lip2vec,laux_2024litevsr,zhang_cvpr_2019}, was to directly predict words or entire sentences based on visual cues. Although simple to implement, these methods face challenges related to speaker-dependent variances, vast vocabularies, and subtle visual differences between words. Recent studies \cite{yeo_aaai_2025,thomas_2025vallr} have utilized phoneme-level prediction to overcome these constraints, viewing lip-reading as a sequence classification problem with a more manageable, smaller number of phoneme classes. Since a language model may be used to reassemble predicted phonemes into words or sentences, phoneme-level modeling can improve data efficiency and generalization. However, a homophone problem arises when two words share the same phoneme sequence (for example, \enquote{see} and \enquote{sea}) but different pronunciations, making it challenging to distinguish sounds solely from visual signals and necessitating context-aware decoding in order to resolve ambiguities.

In this paper, we propose a two-stage, phoneme-centric framework, termed PV-ASR, that combines visual speech features with dense lip landmarks to predict phonemes, which are subsequently mapped to natural language using an encoder-decoder LLM, namely NLLB \cite{costa_2022nllb}. Our primary objective is to evaluate the robustness of the proposed approach under varying lighting conditions and speaker-specific facial variations, while reducing viseme ambiguity through LLM-based sentence reconstruction.

The main contributions of this study are: 1) We propose a fusion of dense spatiotemporal lip landmarks, retaining the lips and chin region with visual features using Spatiotemporal Graph Convolutional Network (ST-GCN) and Convolutional Neural Network (CNN) backbones. This fusion method improves phoneme prediction by leveraging both appearance and articulatory geometry. 2) Instead of relying on a large autoregressive LLM, we employ NLLB, a compact encoder-decoder model, for phoneme-to-text reconstruction. 3) Introducing noise during NLLB training further reduces WER, suggesting increased tolerance to noisy inputs.

\begin{figure*}[t]
  \centering
   \includegraphics[scale=0.048]{figures_draft/ICASSP_lip_reading_architecture_v4.pdf} 

   \caption{Overview of the proposed architecture. The Visual Encoder extracts visual features, while the Landmark Encoder extracts lip landmark features. These two representations are fused and passed through a CTC projection and transformer decoder for sequence modeling. The fused outputs are further processed by the NLLB model to reconstruct phonemes.}
   
   \label{fig:overall_architecture}
\end{figure*}

\section{Methodology}
\label{sec:method}

\noindent
\textbf{Visual Encoder.} Building on prior work \cite{ma_2023avsr}, the first layer of the V-ASR Encoder front-end is a spatiotemporal convolutional layer with a stride of $1 \times 2 \times 2$ and a kernel size of $5 \times 7 \times 7$, which is based on a 3D CNN and a modified ResNet-18. This front-end extracts low- and mid-level spatiotemporal features from the input video frames. The features are then passed to a 12-layer Conformer, which serves as the temporal back-end. The Conformer combines multi-head self-attention and convolutional modules to capture both long-range temporal dependencies (e.g., phoneme coarticulation across frames) and local temporal patterns (e.g., short lip movements). This allows the model to produce context-aware embeddings for each time step.

\noindent
\textbf{Landmark Encoder.} The Landmark Encoder comprises a front-end ST-GCN inspired by \cite{liu_2020stgcn}, followed by a temporal Conformer back-end. The ST-GCN consists of 6 sequential modules, each combining a spatiotemporal graph convolution. The first module takes 2 input channels (from the landmark features) and outputs 64 channels. All subsequent modules take these 64 channels as input and maintain 64 channels as output throughout. The final output is then passed through a linear layer with BatchNorm and a Mish activation to produce the landmark embeddings. The Conformer serves as the temporal back-end, encoding the motion dynamics of lip landmarks into context-aware embeddings. 

\noindent
\textbf{Fusion Layer.} The embeddings from the encoders are passed through a multi-layer perceptron head, which consists of a linear layer followed by BatchNorm, a ReLU activation, and a final linear layer. This head projects the encoder output into a task-specific feature space while introducing non-linearity and stabilizing training through batch normalization.

\noindent
\textbf{Connectionist Temporal Classification (CTC) \cite{watanabe_IEEE_2017}.} The resulting embeddings are projected through a linear layer to obtain phoneme logits, which are used to compute the CTC loss. This loss enforces monotonic alignment between both visual and landmark features and phoneme sequences without requiring frame-level annotations. In the hybrid CTC/attention setup, it complements the decoder loss, accelerating convergence and stabilizing training. 

\noindent
\textbf{Transformer Decoder.} The decoder first maps input tokens into embeddings using an embedding layer combined with positional encoding. These embeddings are then processed by a stack of decoder layers, each containing self-attention, encoder-decoder attention, and position-wise feed-forward sublayers with residual connections and dropout. The decoder embeddings are normalized and projected through a linear layer to the target phoneme space.

\noindent
\textbf{NLLB Transformer Encoder–Decoder \cite{costa_2022nllb}.} The NLLB model is a sequence-to-sequence Transformer where the encoder maps input phoneme sequences to context-aware embeddings using self-attention and feed-forward layers, and the decoder generates words by attending to these embeddings and previously generated tokens. The final decoder outputs are projected through a linear layer and softmax to produce probability distributions over the word vocabulary, enabling sequential reconstruction of complete English sentences from phonemes.

\section{Experimental Setup}
\label{sec:experiment}

\noindent
\textbf{Datasets.}
Experiments are conducted on the LRS2 \cite{son_2017lrs2}, LRS3 \cite{afouras_2018lrs3}, and LRW \cite{chung_2017lrw} publicly available English datasets. LRS2 contains 144k utterances (224.5 h), split into pretraining (96k, 195 h), training (46k, 28 h), validation (1.1k, 0.6 h), and test (1.2k, 0.5 h) subsets. LRS3 has 152k utterances (438.9 h), divided into pretraining (119k, 408 h), train-val (32k, 30 h), and test (1.3k, 0.9 h). LRW is a word-level dataset with TV snippets from the BBC.

\begin{table*}[t]
\centering
\caption{Comparison of phoneme-based lip-reading methods on LRS2 and LRS3 datasets. The table lists the input modality, the LLM used, additional data employed, and total hours, which include both pretraining and training datasets.}
\renewcommand{\arraystretch}{0.95}
\label{tab:sota_lrs2lrs3}
\resizebox{\linewidth}{!}{
\begin{tabular}{lccccccccc}
\toprule
\multirow{2}{*}{\textbf{Method}} & \multirow{2}{*}{\makecell{\textbf{Phoneme} \\ \textbf{-based}}} & \multirow{2}{*}{\textbf{Input}} & \multirow{2}{*}{\textbf{LLM}} & \multirow{2}{*}{\makecell{\textbf{Extra} \\ \textbf{Data}}} & 
\multicolumn{2}{c}{\textbf{LRS2}} & \multicolumn{2}{c}{\textbf{LRS3}} \\
\cmidrule(lr){6-7} \cmidrule(lr){8-9}
 & & & & & \textbf{Total Hours} & \textbf{WER [\%]} & \textbf{Total Hours} & \textbf{WER [\%]} \\
\midrule
ASSTGCN \cite{sheng_2021asstgcn} & - & Video+38-Points & - & - & 223 & 55.7 & 438 & 62.7 \\
Hyb.-Conf. \cite{ma_IEEE_2021} & - & Video & - & \cmark & 381 & 37.9 & 590 & 43.3 \\
VTP \cite{prajwal_2022subword} & - & Video & - & \cmark & 698 & 48.9 & 698 & 40.6 \\
CM-aux \cite{ma_2022multilanguage} & - & Video & - & - & 223 & 32.9 & 438 & 36.3 \\
VTP \cite{prajwal_2022subword} & - & Video & - & \cmark & 2676 & 22.6 & 2676 & 30.7 \\
Auto-AVSR \cite{ma_2023avsr} & - & Video & - & \cmark & 818 & 27.9 & 818 & 33.0 \\
Auto-AVSR \cite{ma_2023avsr} & - & Video & - & \cmark & 3448 & 14.6 & 3448 & 19.1 \\
VALLR \cite{thomas_2025vallr} & \cmark & Video & LLaMA & - & 28 & 20.8 & 30 & 18.7 \\
ViT-3D \cite{serdyuk_2022} & - & Video & - & \cmark & - & - & 90000 & 17.0 \\
Ours(P-ASR) & \cmark & 117-Points & NLLB(1.3B) & - & 223 & 72.2 & 438 & 66.4 \\
Ours(V-ASR) & \cmark & Video & NLLB(1.3B) & - & 223 & 17.1 & 438 & 17.3 \\
Ours(PV-ASR) & \cmark & Video+117-Points & NLLB(1.3B) & - & 223 & 16.0 & 438 & 20.3 \\
\bottomrule
\end{tabular}
}
\end{table*}

\noindent
\textbf{Preprocessing.}
The mouth region is cropped using a $96 \times 96$ pixel bounding box. Each frame is normalized by subtracting the mean and dividing by the standard deviation computed from the training set. Facial landmarks are then extracted using MediaPipe \cite{lugaresi_2019mediapipe}. If the face detection is not successful, zero padding of the landmarks is added to the sequence of landmark frames. We only consider the inner and outer lip landmarks and the surrounding lip landmarks, totaling 117 facial points, including the jaw area. For English Grapheme-to-Phoneme processing, text containing numbers, dates, and currency expressions is first normalized into spoken form using the NVIDIA NeMo toolkit \cite{zhang_2024nemo}. Phoneme sequences are then generated using the open-source SoundChoice toolkit \cite{ploujnikov_2022soundchoice}.

\noindent
\textbf{ARPAbet Phonemes.}
We use the ARPAbet phoneme set in our experiments, a standardized ASCII-based notation commonly used in speech recognition. It consists of 39 English phonemes covering both vowels and consonants, plus two special tokens: \texttt{<unk>} for unknown phonemes and \texttt{\_} for space. The full set is \texttt{AA, AE, AH, AO, AW, AY, B, CH, D, DH, EH, ER, EY, F, G, HH, IH, IY, JH, K, L, M, N, NG, OW, OY, P, R, S, SH, T, TH, UH, UW, V, W, Y, Z, ZH}.

\noindent
\textbf{Stage 1 Training.} 
We fine-tune the P-ASR, V-ASR, and PV-ASR methods on the LRS2 and LRS3 with phoneme labels, respectively. The pretrained weights of the video feature extractor, CTC projector, and transformer decoder are initialized from \cite{ma_2023avsr}, which are pretrained on a 5000-English word vocabulary class from LRW, LRS2, and LRS3. During Stage 1, the visual feature extractor (3D CNN and 2D ResNet-18) is kept frozen to preserve previously learned representations. In contrast, the ST-GCN layers are initialized with LRW-pretrained weights. The remaining layers are randomly initialized. 

\noindent
\textbf{Stage 2 Training.} 
By utilizing readily available text, we train the LLM to learn phonetic–linguistic mappings without relying on visual features. To mitigate phoneme errors propagated from Stage 1, we apply phoneme-level data augmentation during Stage~2 training using random insertions, deletions, and substitutions. This strategy simulates realistic Stage 1 error patterns and encourages reliance on contextual information rather than exact phoneme alignments. Progressive fine-tuning is conducted by first adapting pretrained LLMs on combined LRS2 and LRS3 phoneme data, followed by dataset-specific fine-tuning. Finally, beam search is applied during decoding to select the most plausible sentence output. To ensure fair comparison, all experiments use an identical decoding configuration, including beam search and language model settings, with only the level of phoneme augmentation varied during training.

\noindent
\textbf{Loss Functions.} 
We adopt a hybrid CTC/Attention framework \cite{watanabe_IEEE_2017}, where the training objective is defined as $\mathcal{L} = \alpha \mathcal{L}_{\mathrm{CE}} + (1 - \alpha)\mathcal{L}_{\mathrm{CTC}}$, with $\alpha \in [0,1]$ controlling the relative contribution of the cross-entropy and CTC losses.

\noindent
\textbf{Experiment Setup.} 
We train Stage 1 and Stage 2 models using an NVIDIA A6000 GPU with 49GB of memory. For Stage 1 PV-ASR training, inspired by the experiment protocol in \cite{ma_2023avsr}, we train the model for 50 epochs with the AdamW optimizer, a cosine learning rate scheduler, and a warm-up of 5 epochs. The initial learning rate is $1 \times 10^{-4}$. The maximum number of frames in each training batch is 1,800 frames. For evaluation, we report results by averaging the model checkpoints from the last 10 epochs, which reduces variance and mitigates the effect of fluctuations in individual runs. For Stage 2, we train the LLMs using the AdamW optimizer with an initial learning rate of $5 \times 10^{-5}$. We use the default English tokenizer and treat the phonemes as a special language, which the model learns to reconstruct into natural language text.

\noindent
\textbf{Evaluation Metrics.} 
To evaluate lip-reading performance, we adopt WER \cite{jelinek_1975wer} as our lip-reading evaluation metric. WER accounts for deletions, insertions, and substitutions when calculating the percentage of errors in the predicted text relative to the ground truth. In addition, we calculate the Character Error Rate (CER) to quantify errors at the character level, and Phoneme Error Rate (PER) to evaluate the accuracy of phoneme-level predictions, providing a more fine-grained assessment of lip-reading performance.

\section{Experimental Results}
\label{sec:results}

\noindent
\textbf{Comparison with State-of-the-Art Methods.}
We report the WER on LRS2 and LRS3 in Table \ref{tab:sota_lrs2lrs3}. Most related works pretrain their models on large-scale external datasets and fine-tune on LRS2 and LRS3, respectively, for fair comparison. Our method achieves competitive performance on LRS2 (16.0\% WER), outperforming several existing approaches. Although \cite{ma_2023avsr} reports a slightly lower WER, their model benefits from substantially longer pretraining hours (3448 hours), suggesting that our method is more data-efficient. On the more challenging LRS3 dataset, our method surpasses the majority of recent works. While \cite{thomas_2025vallr} reports 18.7\% WER using a significantly larger LLM architecture, and \cite{serdyuk_2022} achieves 17.0\% WER after training on 90,000 hours of data, our method attains competitive performance with a lighter LLM architecture and using only LRS3 data. One contributing factor is that LRS3 contains more complex scenes, where MediaPipe's facial landmark detection occasionally fails on profile or occluded frames. 

\noindent
\textbf{Comparison of Stage 1 Results for P-ASR, V-ASR, and PV-ASR.}
On the LRS2 dataset, PV-ASR achieves a CER of 5.03\% and a PER of 8.47\%, outperforming both the P-ASR baseline (25.7\% CER, 41.8\% PER) and the V-ASR baseline (5.33\% CER, 9.10\% PER). In contrast, on LRS3, PV-ASR attains a CER of 6.01\% and a PER of 10.09\%, representing slight increases of 0.52\% CER and 0.95\% PER compared to V-ASR (5.49\% CER, 9.14\% PER). Nevertheless, PV-ASR substantially outperforms P-ASR on LRS3, which records a CER of 22.30\% and a PER of 36.96\%. These results indicate that incorporating dense lip landmarks is beneficial on LRS2, while insufficient landmark quality or coverage may slightly degrade performance on LRS3. 

\begin{table}[t]
  \centering
  \caption{WER (\%) of different encoder–decoder LLMs on the LRS2 and LRS3 datasets using greedy and beam search decoding. To evaluate robustness, models were trained with varying levels of introduced errors.}
  \label{tab:wer_da}
  \renewcommand{\arraystretch}{0.9}
  \resizebox{\linewidth}{!}{
  \begin{tabular}{@{}llc|cc|cc@{}}
    \toprule
    \multicolumn{2}{c}{\textbf{Model}} & \multirow{2}{*}{\makecell{\textbf{Noise} \\ \textbf{Err [\%]}}} & \multicolumn{2}{c}{\textbf{LRS2 WER [\%]}} & \multicolumn{2}{c}{\textbf{LRS3 WER [\%]}} \\
    \cmidrule(lr){1-2} \cmidrule(lr){4-5} \cmidrule(lr){6-7}
    Stage 1 & Stage 2 & & Greedy & Beam & Greedy & Beam \\
    \midrule
    \multirow{3}{*}{V-ASR} & \multirow{3}{*}{\makecell{Flan-T5 \\(77M)}} 
      & 0.0 & 31.95 & 30.50 & 44.37 & 33.17 \\
    &  & 10.0 & 26.91 & 25.38 & 40.25 & 26.29 \\
    & & 20.0 & 30.63 & 28.08 & 39.66 & 28.90 \\
    \midrule
    \multirow{3}{*}{PV-ASR} & \multirow{3}{*}{\makecell{Flan-T5 \\(77M)}} 
      & 0.0 & 27.16 & 26.30 & 49.13 & 34.58 \\
    &  & 10.0 & 24.62 & 22.33 & 41.20 & 28.23 \\
    & & 20.0 & 27.73 & 25.59 & 42.15 & 31.51 \\
    \midrule
    \multirow{3}{*}{V-ASR} & \multirow{3}{*}{\makecell{BART \\(139M)}} 
      & 0.0 & 28.77 & 28.60 & 30.75 & 30.46 \\
    & & 10.0 & 19.89 & 19.72 & 19.71 & 19.47 \\
    & & 20.0 & 20.63 & 20.17 & 19.81 & 19.30 \\
    \midrule
    \multirow{3}{*}{PV-ASR} & \multirow{3}{*}{\makecell{BART \\(139M)}} 
      & 0.0 & 29.60 & 24.50 & 33.69 & 33.22 \\
    & & 10.0 & 18.37 & 18.09 & 23.25 & 22.95 \\
    & & 20.0 & 18.46 & 18.21 & 22.67 & 22.43 \\
    \midrule
    \multirow{3}{*}{V-ASR} & \multirow{3}{*}{\makecell{NLLB \\ (600M)}} 
      & 0.0 & 28.96 & 28.03 & 30.71 & 29.99 \\
    &  & 10.0 & 19.74 & 19.02 & 20.08 & 19.22 \\
    & & 20.0 & 19.61 & 18.73 & 19.95 & 18.82 \\
    \midrule
    \multirow{3}{*}{PV-ASR} & \multirow{3}{*}{\makecell{NLLB \\ (600M)}} 
      & 0.0 & 24.52 & 24.14 & 32.50 & 31.87 \\
    &  & 10.0 & 18.15 & 17.84 & 23.92 & 22.87 \\
    & & 20.0 & 18.31 & 17.29 & 23.78 & 21.86 \\
    \midrule  
    \multirow{3}{*}{V-ASR} & \multirow{3}{*}{\makecell{NLLB \\ (1.3B)}} 
      & 0.0 & 39.32 & 28.68 & 31.26 & 30.64 \\
    &  & 10.0 & 18.57 & 17.91 & 18.47 & 17.97 \\
    & & 20.0 & 17.96 & 17.08 & 18.39 & 17.33 \\
    & & 30.0 & 20.18 & 18.76 & 20.02 & 18.23 \\
    
    \midrule    
    \multirow{3}{*}{PV-ASR} & \multirow{3}{*}{\makecell{NLLB \\ (1.3B)}} 
      & 0.0 & 25.27 & 24.93 & 33.26 & 32.90 \\
    &  & 10.0 & 17.20 & 16.48 & 22.84 & 21.82 \\
    & & 20.0 & 17.08 & 16.03 & 21.77 & 20.26 \\
        & & 30.0 & 19.16 & 17.70 & 22.74 & 21.32 \\
    
    \bottomrule
  \end{tabular}
  }
\end{table}

\noindent
WER comparison of V-ASR and PV-ASR combined with four different LLMs, which are Fine-tuned Language Net T5 (Flan-T5) Small (77M), Bidirectional and Auto-Regressive Transformers (BART) Base (139M), NLLB-600M, and NLLB-1.3B models, is shown in Table \ref{tab:wer_da}. To capture fine-grained differences, WER is reported with two-decimal precision in this ablation study. For evaluation, the data augmentation of introducing errors of 0\%, 10\%, and 20\% is tested for all LLMs, and 30\% is conducted only on the final proposed model. Overall, the highest WER across all evaluated LLMs is observed under the 0\% error condition, while performance improves when 10\% or 20\% phoneme errors are introduced during training and inference with beam search decoding, but degrades at 30\% error. Additionally, WER on both LRS2 and LRS3 consistently decreases as the model size of the LLMs increases. Our proposed method of PV-ASR and NLLB-1.3B achieved the best performance on LRS2, 16.0\% WER, and LRS3, 20.3\% WER. Since our results show that NLLB consistently achieved the highest accuracy, likely due to its large-scale multilingual pretraining, which is particularly well-suited for mapping phoneme sequences into coherent text, we select NLLB-1.3B as our final LLM.

\section{Limitations and Future Work}

A limitation of the current study is that large head poses may lead to undetected faces, resulting in missing dense facial landmarks. When fusing the extracted visual and landmark features, this lack of data can hinder the ST-GCN from learning important features effectively. Future research will focus on improving facial landmark detection under challenging conditions, systematically studying the effect of different numbers of dense landmarks and ST-GCN blocks, and developing more advanced fusion strategies, such as attention-based or adaptive gating mechanisms, to better leverage complementary visual and structural cues. In addition, evaluation across different languages and phoneme sets will be explored to assess cross-linguistic generalization. These directions aim to improve the robustness and overall performance of the PV-ASR and NLLB framework.

\section{Conclusions}
This study proposes a two-stage phoneme-based framework that leverages visual features and dense lip landmark motions to predict phonemes, followed by a pretrained NLLB model to reconstruct sentences. By modeling at the phoneme level, we address the inherent ambiguities of viseme-based lip-reading, enabling more accurate speech recognition from visual inputs. Experimental results demonstrate that the proposed method achieves competitive performance on LRS2, while on LRS3, it shows a slightly higher WER. We also highlight the benefits of incorporating lip landmark features in handling variations in lighting and speaker-specific facial characteristics. Moreover, our evaluation results indicate that the chosen NLLB-1.3B model is tolerant to noisy inputs and contributes to improved overall recognition performance.

\section*{Acknowledgments}

This work was supported by JSPS KAKENHI Grant Number JP23H03787.

\vfill\pagebreak

\bibliographystyle{IEEEtran}
\bibliography{paper}

\end{document}